\documentclass{WileyMSP-template}

\usepackage{graphicx}
\usepackage{bm}
\usepackage{amsmath}
\usepackage{amsfonts}
\usepackage{algpseudocode}	
\usepackage{algorithm}
\usepackage{url}
\usepackage{booktabs} 
\usepackage{hyperref}  
\usepackage{multirow}
\usepackage{array}
\usepackage{afterpage}
\usepackage{graphicx}
\usepackage{epstopdf}
\usepackage{setspace}

\begin{document}

\pagestyle{fancy}

\title{Multimodal Fusion for Sim2real Transfer in Visual Reinforcement Learning}

\maketitle

\author{Zichun Xu}
\author{Jingdong Zhao}
\author{Chenyu Guo}
\author{Qianxue Zhang}
\author{Liao Zhang}
\author{Xiao Zhang}
\author{Yiming Ren}
\author{Lian Zhang*}
\author{Zengren Zhao}

\begin{affiliations}
Zichun Xu, Chenyu Guo, Qianxue Zhang, Liao Zhang, Xiao Zhang, Yiming Ren, Lian Zhang, and Zengren Zhao \\
Address: Medical Artificial Intelligence Lab, The First Hospital of Hebei Medical University, Hebei Medical University, Shijiazhuang 050000, China. \\
Email Address: lianzhang@hebmu.edu.cn

Jingdong Zhao \\
Address: State Key Laboratory of Robotics and Systems, Harbin Institute of Technology, Harbin 150001, Heilongjiang Province, China.

\end{affiliations}


\begin{abstract}
	Depth information is robust to scene appearance variations and inherently carries 3D spatial details. Thus, a visual backbone based on the vision transformer is proposed to fuse RGB and depth modalities for enhancing generalization in this paper. Different modalities are first processed by separate CNN stems, and the combined convolutional features are delivered to the scalable vision transformer to obtain visual representations. Moreover, a contrastive learning scheme is designed with masked and unmasked tokens to enhance the sample efficiency and generalization performance. A curriculum-based domain randomization scheme is used to flexibly stabilize the training process. Finally, simulation results demonstrate that our fusion scheme outperforms the other baselines. The feasibility of our model is validated to perform real-world manipulation tasks via zero-shot transfer. 
\end{abstract}

\justifying

\section{INTRODUCTION}
Reinforcement learning (RL) has exhibited its superior ability in addressing contact-rich tasks without a tedious dynamics model. Recent works focus on integrating different modality information tailored to specific task scenarios, in which vision \cite{chen2024multimodality}, proprioception \cite{noh2022toward}, force/torque \cite{jin2024visualforcetactile}, and tactile \cite{hansen2022visuotactilerl} are prevalent alternatives. Considering real-world situations and the cost of acquiring information, vision is gradually emerging as an essential modality. In contrast to state-based data, high-dimensional vision streams need to be encoded as compact representations via visual backbones. The low sample efficiency of pixel-based RL has motivated numerous studies dedicated to enhancing its performance \cite{xu2024drm}.

High training costs in visual RL typically necessitate simulation-to-reality (sim2real) transfer or real-world fine-tuning \cite{julian2020never, ze2023visual}. The sim2real transfer performance is closely related to the generalizability of the visual encoder. However, generalizability is a relatively ambiguous metric for evaluations. Some works suggest that a vision backbone capable of extracting object- or agent-centric representations performs better in diverse environments, benefiting from its weak sensitivity to irrelevant distractors \cite{gmelin2023efficient, pore2024dear}. Generally, domain randomization is employed to narrow the gap between simulation and reality in previous works \cite{peng2018simtoreal, josifovski2022analysis}, which are limited to boost the transfer performance. For the further exploitation of visual information, a visual backbone pretrained on in-the-wild datasets has been demonstrated to be effective in extracting generalized representations across different conditions. However, representation learning focuses on acquiring general and transferable visual representations that are independent of specific task requirements \cite{radosavovic2022realworld}. Extra training is necessary to perform policy optimization and increase long-term returns. Hence, the perception-then-decision joint optimization procedure of visual RL prevails as the popular training paradigm. Furthermore, other modalities can be pursued for additional gains.

\begin{figure}[t!]
	\centering
	\includegraphics[scale=.35]{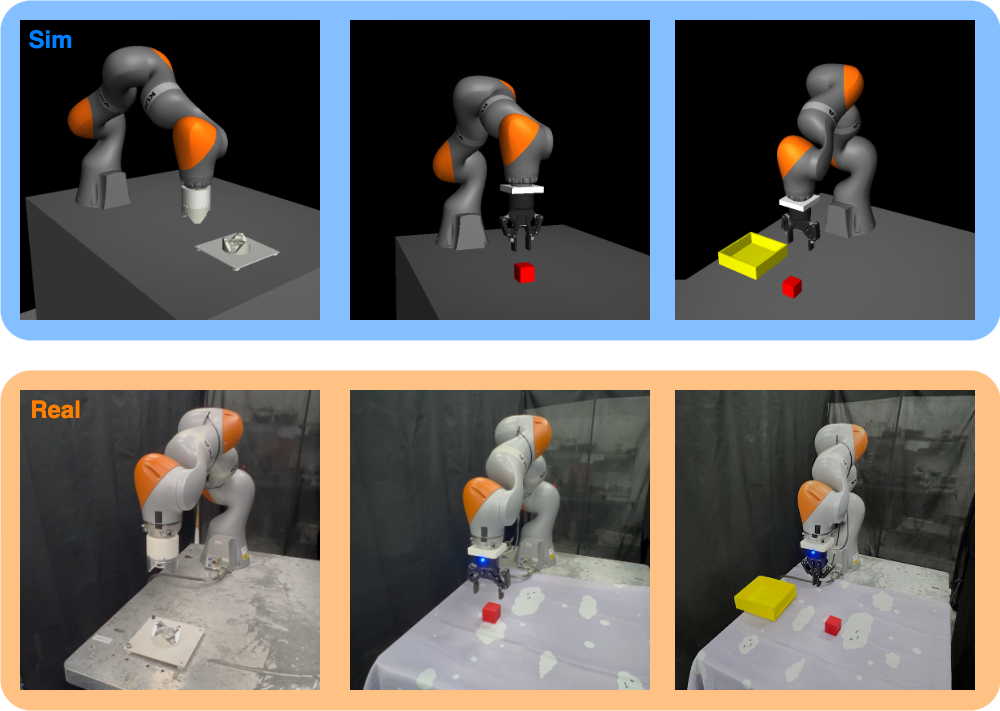}
	\caption{Zero-shot transfer to complete manipulation tasks.}
	\label{intro}
\end{figure}

Compared to RGB images, depth information can better mitigate perceptual disparities and provide distance information to empower spatial awareness. Depth images enable less training burden than other forms of representations, e.g., point clouds. A common utilization of depth images is to serve as an extra channel of RGB. Empirical evidence from computer vision demonstrates that the fusion strategy and framework for different modalities play a pivotal role in semantic segmentation or detection \cite{zhang2022spatio, zhang2023cmx}, revealing that the power of depth information can be further unleashed through reasonable vision encoders \cite{zhang2023cmx}. Nonetheless, the primary RL research focuses on exploring different training paradigms to enhance visual generalization \cite{hansen2021stabilizing, bertoin2022look}. The visual backbone architecture can also influence the performance of modality fusion. Inspired by the above, this paper introduces a deeper fusion paradigm with the self-attention mechanism of the vision transformer (ViT) \cite{dosovitskiy2021image}. Meanwhile, a contrastive unsupervised learning scheme tailored for ViT is proposed to enhance sample efficiency. Extensive simulation experiments demonstrate that the proposed vision backbone can better utilize the depth information and thus improve the generalization. Last, a curriculum-based domain randomization is refined to stabilize training processes for zero-shot sim2real transfer, which is shown in Figure~\ref{intro}. In summary, our contributions are as follows: 

\begin{enumerate}
	\item We propose a visual backbone based on ViT and CNN to fuse RGB-D information. The trained vision backbone demonstrates the capability to extract object- and agent-centric representations while effectively exploiting the complementary advantages of each visual modality.
	\item The contrastive learning scheme is configured with masked and unmasked convolutional features to improve sample efficiency and policy generalization.
	\item Extensive ablation experiments are performed to evaluate the impact of the proposed contrastive learning scheme and key architectural components on generalization performance.
	\item The visual backbone and policy network trained through curriculum-based domain randomization can be transferred to the real world in a zero-shot manner.
\end{enumerate}

\section{RELATED WORK}
\subsection{Visual RL}
Visual RL is an intersection field spanning RL and computer vision. A prevalent framework is to integrate the training of the visual backbone with the policy network, which facilitates the extraction of in-domain features \cite{bertoin2022look, yuan2023rlvigen}. For this training paradigm, limited visual diversity and unlabeled data highlight the importance of improving sample efficiency. Data augmentations \cite{laskin2020reinforcement}, such as random shift and color jitter, can significantly enhance the sample efficiency and adaptability of the agent to unseen environments. Extra recipes are essential for stabilizing critic updates during training. Denis et al. propose the renowned DrQ \cite{yarats2021image} and its successor, DrQ-v2 \cite{yarats2022mastering}, which derive the critic target from weak-augmented visual observations. In contrast, SVEA \cite{hansen2021stabilizing} allows weakly and strongly augmented observations to share the same critic target. In addition to meticulous recipes, the visual backbone is also relevant to sample efficiency and visual attention regions. A CNN-based visual backbone is preferred due to its lightweight architecture compared to ViT. Some well-established vision architectures, such as ResNet18 \cite{he2016deep} and ViT-B \cite{dosovitskiy2021image}, are oversized for online RL despite being highly acclaimed in computer vision. Online RL requires frequent data interactions to continuously update the policy, and some large-scale model architectures, even pre-trained versions, are time-consuming \cite{xiao2022masked, parisi2022unsurprising}. A scalable ViT and the first two layers of the pre-trained ResNet18 have been validated as visual backbones for training off-policy algorithms \cite{hansen2021stabilizing, yuan2022pretrained}. Therefore, for our work, a scaled ViT with two CNN stems is constructed as the visual backbone to fuse RGB and depth information for online training.

\subsection{Generalization in Visuomotor Control}
RGB has become an indispensable component when training with the visual modality \cite{chen2024multimodality, jangir2022look}. Like human binocular perception, RGB images enable faster comprehension of complex scenes through scene changes. The sim2real discrepancy persists despite photorealistic rendering and necessary domain randomization in some simulators. A robust visual backbone with strong generalization has a crucial impact on the sim2real transfer. Specifically, an ideal visual backbone focuses on task-relevant regions. Manipulation centricity \cite{jiang2024robots} is proposed as an indicator to quantify the performance of trained visual backbones for downstream manipulation tasks. Pre-training on robot-related datasets \cite{khazatsky2024droid} can better reduce the domain gap compared to those on daily-life datasets. Similarly, related works attempt to obtain disentangled or agent-related representations that perform robustly against distracting backgrounds \cite{gmelin2023efficient, pore2024dear}. However, some sim2real works primarily treat RGB as the main modality while rarely utilizing depth information \cite{yuan2022simtoreal, tobin2017domain}. The embedded distance information and the insensitivity to appearance variations are valuable properties of depth information for enhancing visual generalization. The injection of random noise during simulation followed by a denoising process is sufficient without elaborate domain randomization for depth information \cite{yuan2024learning}. For applications in RL, different modalities are typically integrated via channel concatenation, which is described as early fusion \cite{lee2025grasping}. In this paper, we move beyond simple early fusion via channel concatenation. Instead, the self-attention mechanism of ViT is used to deeply exploit the depth modality.

\section{Preliminaries}
\subsection{Visual RL} 
The visual RL process can be formulated as a partially observable Markov decision process (POMDP), defined as a tuple $ \langle \mathcal{O}, \mathcal{A}, \mathcal{P}, \mathcal{R}, \rho , \gamma \rangle $ with the visual observation space $\mathcal{O}$, action space $\mathcal{A}$, transition function $\bm{o}_{t+1}=\mathcal{P}\left(\cdot |\bm{o}_t, \bm{a}_t\right) $, reward function $r_t = \mathcal{R} \left( \bm{o}_t, \bm{a}_t \right) $, distribution of the initial state $ \bm{o}_0 \sim \rho $, and discount factor $\gamma \in [0, 1)$. A stack of recent frames is adopted as the current observation $\bm{o}_t = \left(\bm{x}_{t-i}, \ldots , \bm{x}_{t}\right) $ to capture temporal information. Our goal is to train policies that maximize the cumulative reward $ \mathbb{E}_{\pi}\left[\sum_{t = 0}^{\infty} {\gamma}^{t}r_t  \right] $ and complete the corresponding task.

\subsection{Vision Transformer} 
ViT is a pioneering application that extends the self-attention mechanism from natural language processing to computer vision. An image $\bm{o}_t \in \mathbb{R}^{H \times W \times C}$ is split into $N = HW/P^2$ non-overlapping patches $\bm{p}_n \in \mathbb{R}^{D}$ and flattened into a linear sequence, where $D = P^2 C$. The CLS token is prepended in the patch sequence, and the positional embeddings $\bm{E}_{pos} \in \mathbb{R}^{\left(N+1\right)  \times D}$ are added for the preservation of the positional information 
\begin{equation}
	\bm{z}_{e}^0 = [\bm{T}_{\text{CLS}}; \bm{p}_1; \cdots ; \bm{p}_N] + \bm{E}_{pos}.
\end{equation}

The transformer encoder consists of $L$ identical layers, each of which includes the multi-head self-attention (MSA) and feed-forward (FFN) blocks. The input vector $\bm{z}_e^{i-1} \in \mathbb{R}^{(N+1) \times D}, i = 1 \ldots L $ to MSA is first linearly projected into query, key, and value, i.e., $\bm{Q}, \bm{K}, \bm{V} \in \mathbb{R}^{h \times (N+1) \times D_h}$. The attention score is formulated as
\begin{equation}
	\bm{O}_{ij} = \text{Softmax} \left(\frac{\bm{Q}_{ij}\bm{K}_{ij}^T}{\sqrt{D_h}}\right) \cdot \bm{V}_{ij} \quad  j = 1 \cdots h
\end{equation}
and added with $\bm{z}_e^{i-1}$ through the residual connection, where $D_h = D/h$ represents the dimension of each head. The entire encoding process with FFN can be summarized as 
\begin{equation}
	\left\{
	\begin{alignedat}{3}
	\bm{z}_e^i = &\ \text{MSA} \left(\text{LayerNorm}\left(\bm{z}_e^{i-1}\right) \right) + \bm{z}_e^{i-1} \\
	\bm{z}_e^i = &\ \text{FFN} \left(\text{LayerNorm}\left(\bm{z}_e^i \right) \right) + \bm{z}_e^i \\
	\bm{z}_e^L = &\ \text{LayerNorm} \left(\bm{z}_e^L \right).
	\end{alignedat}
	\right.
\end{equation} 

Following the FFN block in the final layer, the CLS token of $\bm{z}_e^L$ can be treated as the visual representation $\bm{z}_t$ for downstream RL \cite{hansen2021stabilizing}. The Masked Autoencoder (MAE) \cite{he2022masked} is a self-supervised learning framework that builds upon ViT. A subset of patches is randomly masked, and the visible patches are utilized to construct the latent, which is then employed by the decoder to reconstruct the original image.


\begin{figure*}[t!]
	\centering
	\includegraphics[width=1.\textwidth]{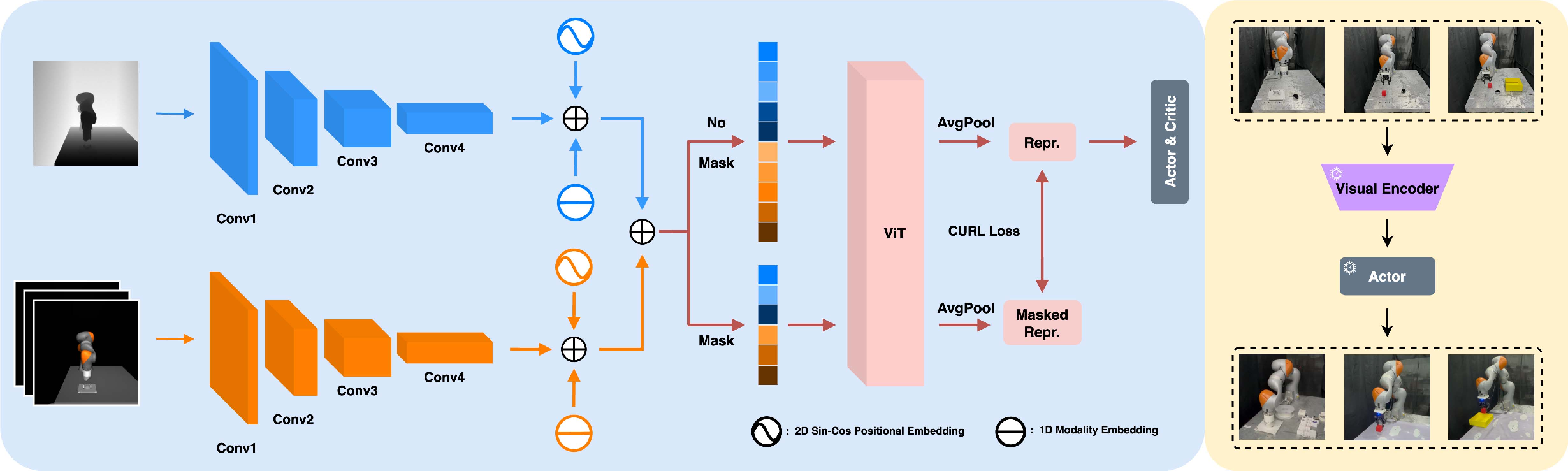}
	\caption{Overview of our approach. We build a visual backbone with two CNN stems to process RGB and depth images, respectively. The convoluted features are concatenated and then passed to ViT to obtain encoded tokens. Average pooling is applied to the unmasked tokens to derive compact visual embeddings for the policy head. Concurrently, contrastive training is performed using masked and unmasked tokens. Trained visual and policy networks are frozen and transferred to reality without fine-tuning.}
	\label{overview}
\end{figure*}

\section{Method}
This section presents a visual backbone based on ViT to fuse RGB and depth images for visual RL. The proposed visual backbone can be integrated into the existing off-policy RL framework to extract representations that encode multimodal visual information. To further improve the generalization performance and the perception of task-related regions, a contrastive objective is formulated to maximize the mutual information between the complete and visible latents after random masking and encoding by ViT. Details will be elaborated in the following sections.

\subsection{Multimodal Visual Encoder}
While ViT is widely employed to encode RGB images, handling depth information requires specific architectural considerations. Following \cite{sferrazza2024power}, we incorporate a dedicated convolutional stem to process the stacked depth observations, as illustrated in Figure~\ref{overview}. Furthermore, since prior studies have indicated that incorporating early convolutions prior to the Transformer layers helps to extract fine-grained details, both RGB and depth images are processed separately via respective convolutional stems:
\begin{equation}
	\bm{h}_t^{rgb} = g_{conv}^{rgb}\left(\bm{o}_t^{rgb}\right)  \qquad  \bm{h}_t^{d} = g_{conv}^{d}\left(\bm{o}_t^{d}\right).
\end{equation}
Individual 2D sin-cos positional and 1D modality embeddings are added to the convolutional features, respectively. Embedding vectors of different modalities are concatenated and passed to ViT $\left(g_{v}\right)$. The visual representation can be obtained by the average pooling over the encoded tokens without an additional MLP head:
\begin{equation}
	\bm{z}_t = g_{v}\left( \bm{h}_t \right) \quad \bm{h}_t = \left[\bm{h}_t^{rgb}; \bm{h}_t^{d}\right].
\end{equation}
The encoded tokens can also be extracted after random masking and then transmitted to an asymmetric decoder to reconstruct the unmasked images of different modalities, which is the implementation of MAE. While a similar study \cite{sferrazza2024power} focuses on reconstructing RGB and tactile observations, the effect of the asymmetric decoder on generalization will be specifically evaluated with more challenging scenarios in Section~\ref{sec_ablation}.

\subsection{Contrastive Learning}
\textbf{C}ontrastive \textbf{U}nsupervised Representations for \textbf{R}einforcement \textbf{L}earning (CURL) \cite{laskin2020curl} aims to maximize the mutual information between the anchor and positive. Inspired by CURL and MAE, masked and unmasked patches naturally serve as the anchor and positive pairs. Unlike the original MAE, random masking is performed on the convolutional features instead of normalized images and performed uniformly across both visual modalities. Random shift augmentation \cite{yarats2022mastering} is employed to augment the observation before masking. Random shuffling is omitted since there is no asymmetric decoder. Note that random masking is exclusively utilized to the CURL phase. The visual representation employed by the actor and critic is unmasked. The loss function of CURL can be represented as
\begin{equation}
	\mathcal{L}_{curl} = - \alpha \log \frac{\exp\left(q^T \cdot k^+/ \tau\right)}{\exp \left( q^T \cdot k^+  / \tau \right) + \sum_- \exp \left(q^T \cdot k^- / \tau \right)},
	\label{curl}
\end{equation}
where $\tau$ is the temperature coefficient. $q = \bm{z}_t$ and $k^+ = g_{v}\left( \bm{h}_t^{mask} \right)$ are query and positive samples, where $\bm{h}_t^{mask} \sim p^{mask} \left(\bm{h}_t^{mask} \vert \bm{h}_t, m \right)$. $k^-$ denotes the negative samples from different timesteps. $\alpha$ is the contrastive loss weight, and $m$ is the mask ratio. In contrast to the original CURL \cite{laskin2020curl}, the query-key pair is generated not by the target network for the visual encoder but instead by the recent weights.

\subsection{Reinforcement Learning Backbone}

The proposed visual encoder and contrastive learning framework are integrated into SVEA \cite{hansen2021stabilizing} to transfer the trained visual backbone and policy network in a zero-shot manner. SVEA is an extension of DrQ-v2 \cite{yarats2022mastering}, in which the weak augmentation recipe is replaced with two separate data streams to further improve visual generalization and stabilize the training process. The strongly augmented observations used in this paper are obtained by randomly overlaying out-of-domain images \cite{zhou2017places} to further improve visual generalization. Given the batch data $\mathcal{B} = \left(\bm{o}_t, \bm{a}_t, r_{t:t+n-1}, \bm{o}_{t+n}\right)$ from the replay buffer $\mathcal{D}$, the actor is updated with the following loss:
\begin{equation}
	\mathcal{L}_\pi = -\mathbb{E}_{\mathcal{B} \sim \mathcal{D} } \bigg[ \min_{k=1,2} Q_{\theta_k} \bigl(f_\xi \left(\bm{o}_t\right), \bm{\tilde{a}}_t \bigr) \bigg],  
	\label{actor_loss}
\end{equation}
where $f_\xi$ indicates the vision encoder. $\bm{\tilde{a}}_t = \pi_\phi \left( f_\xi \left(\bm{o}_t\right) \right) + \epsilon $ is inferred by the policy network only with the weakly augmented stream. $\epsilon$ is the exploration noise. $\bm{o}_t^{\text{aug}}$ refers to the observation after the strong augmentation. The critic loss is defined with both weakly and strongly augmented streams:
\begin{equation}
	\label{critic_loss}
	\mathcal{L}_Q = \mathbb{E}_{\mathcal{B}  \sim \mathcal{D} } \biggl[ \beta _1 \left\lVert  Q_{\theta_k} \left(f_\xi \left(\bm{o}_t\right), \bm{a}_t\right) - y_t^{tgt} \right\rVert _2  ^2 + \beta _2 \left\lVert  Q_{\theta_k} \left(f_\xi \left(\bm{o}_t^{\text{aug}}\right), \bm{a}_t\right) - y_t^{tgt} \right\rVert _2  ^2  \biggr] \quad  \forall k \in \left\{1,2\right\},
\end{equation}
where
\begin{equation}
	y^{tgt} = \sum_{i = 0}^{n-1} \gamma^i r_{t+i} + \gamma^n \bigg( \underset{k=1,2}{\rm{min}} Q_{\bar{\theta}_{k}} \left(f_\xi \left(\bm{o}_{t+n}\right), \bm{\tilde{a}}_{t+n} \right) \bigg)
	\label{critic_target}
\end{equation}
is the critic target obtained only with unaugmented observation. $\beta _1 = \beta _2 = 0.5$ are constants utilized to balance the two data streams. To reduce computational overhead, observations from two distinct data streams can be concatenated along batch dimension. 



\subsection{Curriculum-based Domain Randomization}
Domain randomization is incorporated into the training process to reduce the sim2real gap, typically encompassing variations in scene appearance, lighting conditions, dynamic parameters, and camera viewpoints. Large randomized magnitude will trigger divergence in the RL process. Thus, we implement a curriculum-based domain randomization strategy inspired by \cite{yuan2024learning} to progressively broaden the random range. Upon the initiation of domain randomization, the random ranges of all parameters expand exponentially with increasing episodes:
\begin{equation}
	R_c = R_d \cdot \lambda  + R_r \cdot \left(1 - \lambda^{T_e} \right), 
	\label{domain_rnd}
\end{equation}
where $R_c$ and $R_d$ are randomized and default domain parameters, respectively. $R_r$ represents the predefined random range. $\lambda$ is the decay coefficient over running episodes $T_e$. While \cite{yuan2024learning} relies on frame-based thresholds, the evaluation success rate is employed as a flexible trigger to prevent instability and avoid unnecessary computation caused by premature or delayed activation. Algorithm~\ref{algo} provides a detailed explanation about the training process.

\begin{algorithm}[t]
	\setstretch{1.2}
	\caption{Training details}\label{algo}
	\begin{algorithmic}[1]
		\Require Actor $\pi_\phi$, Critic $Q_{\theta_k}$, visual encoder $f_\xi$, learning rate $\eta$, replay buffer $\mathcal{D}$, training steps $T$, evaluation period $E$, mask ratio $m$, contrastive loss weight $\alpha$, decay coefficient $\lambda$, domain randomization flag $p_{f}$, and CURL period $F$.
		\State $\bm{s}_0 \leftarrow \rho$
		\For {timesteps $t = 1 \ldots T$}
			\State $\bm{\tilde{a}}_t \leftarrow \pi_\phi \left(\cdot | \bm{o}_t\right) + \epsilon$
			\State $\bm{o}_{t+1} \leftarrow \mathcal{P}\left(\cdot |\bm{o}_t, \bm{\tilde{a}}_t\right)$
			\State $\mathcal{D}  \leftarrow \mathcal{D} \cup \left(\bm{o}_t, \bm{\tilde{a}}_t, r_t, \bm{o}_{t+1}\right) $
			\State $\mathcal{B} \leftarrow \mathcal{D}$ \hfill $\qquad \vartriangleright \text{Sample batch transitions}$
			\State \textsc{UpdateCritic}$\left(\mathcal{B}\right)$ with Equation~\ref{critic_loss}
			\State \textsc{UpdateActor}$\left(\mathcal{B}\right)$ with Equation~\ref{actor_loss}
			\If{$t\ \% \ F == 0$}
				\State \text{CURL} \text{with Equation~\ref{curl}} \hfill $\vartriangleright$ \text{Update Visual Encoder}
			\EndIf
			\If{$t\ \% \ E == 0$}
				\State $p_{eval} \leftarrow$ Evaluations
				\If{$p_{eval} \geq p_{f}$}
					\State Domain randomization enabled
				\EndIf
			\EndIf
		\EndFor
	\end{algorithmic}
\end{algorithm}

\section{Simulations}\label{simulations}
This section begins with presenting manipulation tasks and simulation settings used for subsequent testing. Next, different fusion methods for RGB and depth are tested and quantitatively analyzed on the above tasks to serve as references for subsequent sim2real transfer. Last, ablation experiments are finally performed to verify the effect of some potential factors.

\subsection{Tasks}
We develop three manipulation tasks powered by MuJoCo as benchmarks for evaluations, which are shown in Figure~\ref{tasks}. For each task, the KUKA LBR IIWA 14 R820 manipulator is adopted as the main embodiment, and stage-dense rewards are customized. Object position is randomized during episode resets. The details of all tasks are as follows:

\begin{figure}[t!]
	\centering
	\includegraphics[scale=.25]{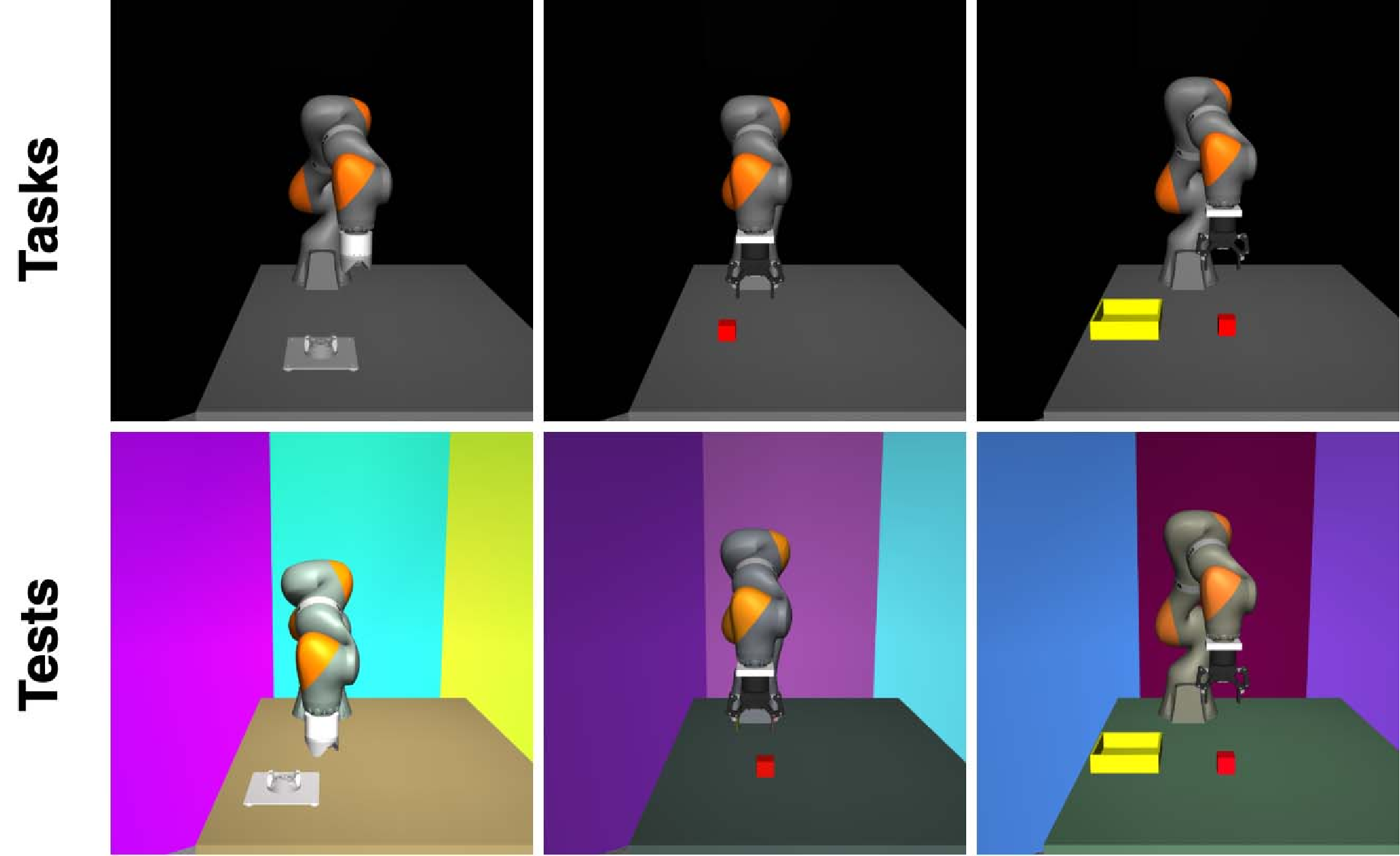}
	\caption{Benchmark environments (top) and unseen scenarios (bottom) for agents during testing, in which tasks are labeled from left to right as Assembly, Lift, and PickAndPlace.}
	\label{tasks}
\end{figure}

\textbf{Assembly:} Modular connectors mounted on the end-effector and table are genderless and serve for on-orbit assembly \cite{10985879}. The collision meshes of connectors are manually segmented before import to optimally maintain their surface contours. This task is considered successful when both connectors achieve the mating state within a tolerance of 1cm. The agent is endowed with 3-DOF motion in the Cartesian space. Due to the presence of numerous contact-rich states during assembly, we extend the action space with variable impedance gains \cite{Martin2019Variable}.

\textbf{Lift:} The Robotiq 2F-85 gripper is attached to the end-effector to grasp the red cube. The action space also contains 3-DOF motion in the Cartesian space with an additional continuous control of the gripper. The red cube should be lifted above the table.

\textbf{PickAndPlace:} This task has the same action space as Lift. Episodes are determined to be successful by picking the red cube first and then placing it in the yellow box.

\subsection{Simulation Results}

This section compares our approach with different fusion approaches: (i) \textbf{Early fusion (EF)} is implemented via channel-wise concatenation in a CNN-based encoder, followed by ReLU activation \cite{yarats2022mastering}. (ii) \textbf{Late fusion (LF)} employs dual CNN stems to process each modality separately. Different encoded representations are flattened and then concatenated before being passed to the policy head. (iii) The pure \textbf{ViT} serves as the visual backbone without any CNN stems. Images of different modalities are partitioned into patches through the standard patchification layer and then concatenated along the patch dimension. (iv) \textbf{PIE-G} \cite{yuan2022pretrained} integrates latent flow processing and the first two layers of the frozen ResNet18 model pre-trained with ImageNet. Pre-trained visual models have been explored to achieve better generalization. Compared to some models with numerous parameters \cite{xiao2022masked}, PIE-G incurs lower wall-clock training overhead and is employed as a representative baseline for comparison in this paper.

A maximum depth threshold is set to normalize the depth image and remap to the pixel range [0, 255]. Additional Gaussian and depth-dependent noise is added and subsequently smoothed with GaussianBlur \cite{yuan2024learning}. The visual observation consists of $3$ RGB frames and $1$ depth frame, each with a size of $128 \times 128$. Our visual backbone includes a 4-layer ViT with 8 heads and 2 convolutional stems, each comprising a 4-layer CNN. All hyperparameters for training are detailed in Table~\ref{common_hyper} in Appendix. The training below is performed on a workstation equipped with an i9-12900KF CPU and two NVIDIA RTX 3090 GPUs.

\begin{figure*}[t!]
	\centering
	\includegraphics[width=1\textwidth]{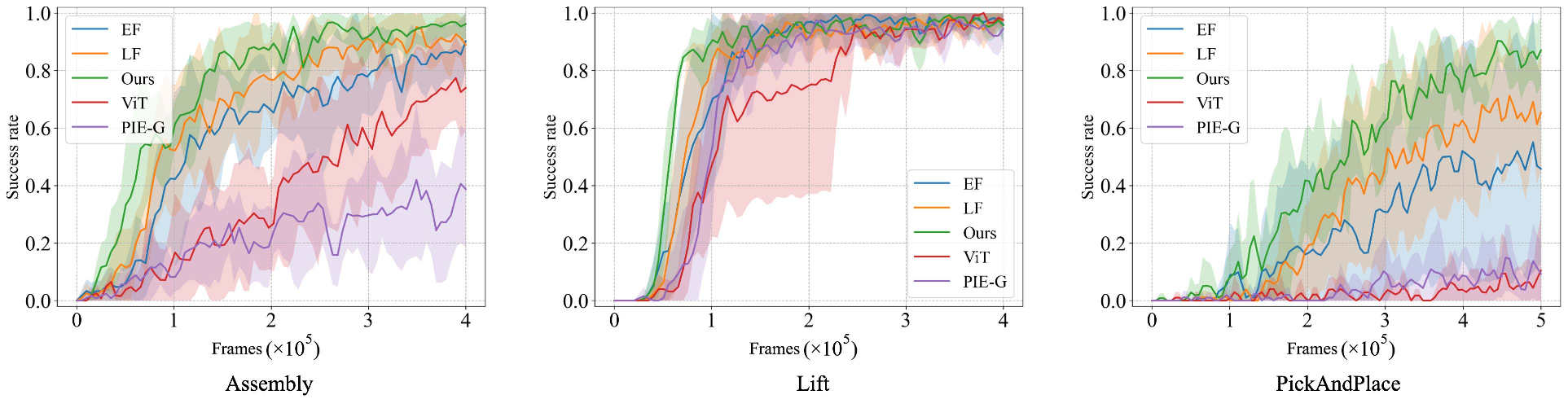}
	\caption{Training curves using different fusion approaches, where solid lines and shaded areas represent the mean and confidence interval over 5 random seeds, respectively.}
	\label{sim_results}
\end{figure*}

\begin{table}[t!]
	\centering
	\caption{Generalization performance with domain randomization disabled and limited training frames. Success rate (mean $\pm$ standard deviation) is evaluated over $100$ random episodes.}
	\label{test_result}
	\begin{tabular}{@{}>{\centering\arraybackslash}m{1.2cm}ccc@{}}
	\toprule  
	Method  & Assembly [\%] & Lift [\%] & PickAndPlace [\%]  \\ 
	\midrule
	\textbf{Ours} & $\bm{82.8 \pm 1.8}$ & $\bm{39.8 \pm 1.7}$ & $\bm{42.0 \pm 4.4}$ \\
	\addlinespace[0.2em]
	EF & $61.8 \pm 13.1$ & $14.8 \pm 6.1$ & $5.4 \pm 4.5$ \\
	\addlinespace[0.2em]
	LF & $77.4 \pm 8.6$ & $26.0 \pm 10.5$ & $14 \pm 7.4$ \\
	\addlinespace[0.2em]
	ViT & $45.4 \pm 17.6$ & $12.0 \pm 6.0$ & $0.2 \pm 0.4$ \\
	\addlinespace[0.2em]
	PIE-G & $2.0 \pm 0.9$ & $ 0.8 \pm 0.9 $ & $ 0.4 \pm 0.5$ \\
	\addlinespace[-0.2em]
	\bottomrule   
	\end{tabular}
\end{table}

\begin{table}[t!]
	\caption{Score-CAM visualization of different visual backbones on all tasks.}
	\centering
	\label{grad_cam}
	\begin{tabular}{	
	>{\centering\arraybackslash}m{1cm}
    >{\centering\arraybackslash}m{1.8cm}
    >{\centering\arraybackslash}m{1.8cm}
    >{\centering\arraybackslash}m{1.8cm}}
	\toprule  
	Task & \includegraphics[width=1.8cm]{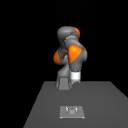} & \includegraphics[width=1.8cm]{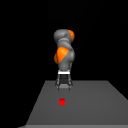} & \includegraphics[width=1.8cm]{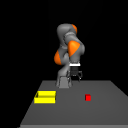}\\ 
	\midrule
	\textbf{Ours} & \includegraphics[width=1.8cm]{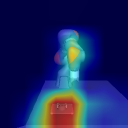} &  \includegraphics[width=1.8cm]{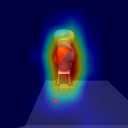} & \includegraphics[width=1.8cm]{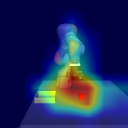}\\
	EF & \includegraphics[width=1.8cm]{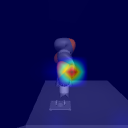} & \includegraphics[width=1.8cm]{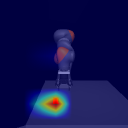} & \includegraphics[width=1.8cm]{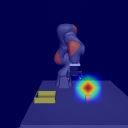} \\
	LF & \includegraphics[width=1.8cm]{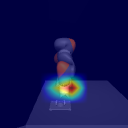}  & \includegraphics[width=1.8cm]{lift_cam_ef.png} & \includegraphics[width=1.8cm]{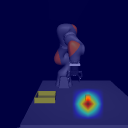} \\
	ViT & \includegraphics[width=1.8cm]{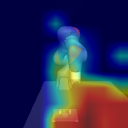} & \includegraphics[width=1.8cm]{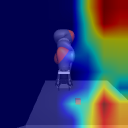} & \includegraphics[width=1.8cm]{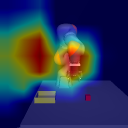} \\
	PIE-G & \includegraphics[width=1.8cm]{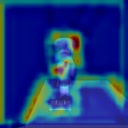} & \includegraphics[width=1.8cm]{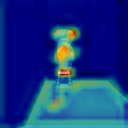} & \includegraphics[width=1.8cm]{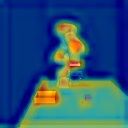} \\
	\bottomrule   
	\end{tabular}
\end{table}

\begin{figure}[t!]
	\centering
	\includegraphics[width=.45\textwidth]{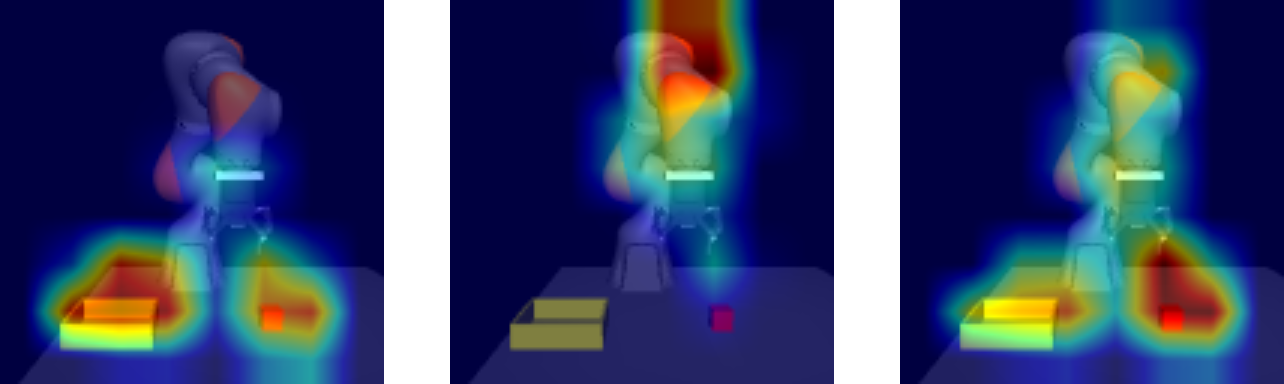}
	\caption{Attention visualization of RGB (left), depth (middle), and RGB-D (right) with the proposed visual backbone.}
	\label{different_attentions}
\end{figure}

Five independent runs are performed with different seeds for each method. To save computational resources and observe the effect of modality fusion methods on generalization, each model is trained over a restricted number of frames and without domain randomization. All training processes are shown in Figure~\ref{sim_results}, in which our method demonstrates better sample efficiency, particularly in the PickAndPlace task. Generalization tests are conducted in the unseen scenarios illustrated in Figure~\ref{tasks}. The test results in Table~\ref{test_result} are obtained using the weights trained as described above. Meanwhile, Score-CAM \cite{wang2020score} visualizations are presented in Table~\ref{grad_cam} to analyze the underlying effects of visual backbones on generalization performance due to the gradient-free frozen weights of PIE-G. Compared to baselines, our visual backbone shows broader attention on agent- and object-related regions. More task-centric areas can be captured and extracted as informative representations for downstream inference, which contributes to the improved performance of our method in Table~\ref{test_result}. Meanwhile, the complementary properties of two modalities are exploited by our visual backbone to focus on different task-related regions, as illustrated in Figure~\ref{different_attentions}. The poor performance of ViT demonstrates that early convolutions can help ViT understand low-level details \cite{xiao2021early} and thus accelerate the convergence of RL processes. For EF and LF, the dual-CNN backbone outperforms channel concatenation for multimodal feature fusion. PIE-G exhibits the lowest performance despite incorporating a pre-trained visual model. As shown in Table~\ref{grad_cam}, PIE-G extracts general features within the field of view, which lacks granularity and task awareness. RL policies tend to overfit to task-irrelevant visual information and thus suffer from reduced generalization. Furthermore, ImageNet lacks images of manipulated objects and robots.

\subsection{Ablations}\label{sec_ablation}
The following ablation experiments are performed on the Lift task under the condition of 400K training frames.

\textbf{CURL.} The contribution of CURL to generalization is evaluated via ablation studies. Equation~\ref{curl} is disabled, and the visual backbone is updated solely via Equation~\ref{critic_loss}. The success rate on the Lift task decreased by approximately $15\%$ in Table~\ref{ablations}, demonstrating that the proposed CURL scheme significantly improves the generalization capabilities of the encoder.

\begin{table}[th]
	\centering
	\caption{Ablation experiments on the Lift task.}
	\label{ablations}
	\begin{tabular}{@{}>{\centering\arraybackslash}m{5cm}c@{}}
	\toprule  
	Ablation  & Success rate [\%] \\ 
	\midrule
	w/ CURL (mask ratio = 0.1) &  $39.8 \pm 1.7$ \\
	\addlinespace[0.2em]
	w/o CURL  & $24.6 \pm 4.5$ \\ 
	\addlinespace[0.2em]
	w/ decoder \& w/o CURL &  $31.8 \pm 6.1$ \\
	\addlinespace[0.2em]
	mask ratio = 0.0 &  $34.2 \pm 3.1$ \\
	\addlinespace[0.2em]
	mask ratio = 0.4 &  $19.8 \pm 8.1$ \\
	\addlinespace[0.2em]
	mask ratio = 0.7 &  $15.8 \pm 7.3$ \\
	\addlinespace[-0.2em]
	\bottomrule   
	\end{tabular}
\end{table}

\begin{figure}[t!]
	\centering
	\includegraphics[width=.6\textwidth]{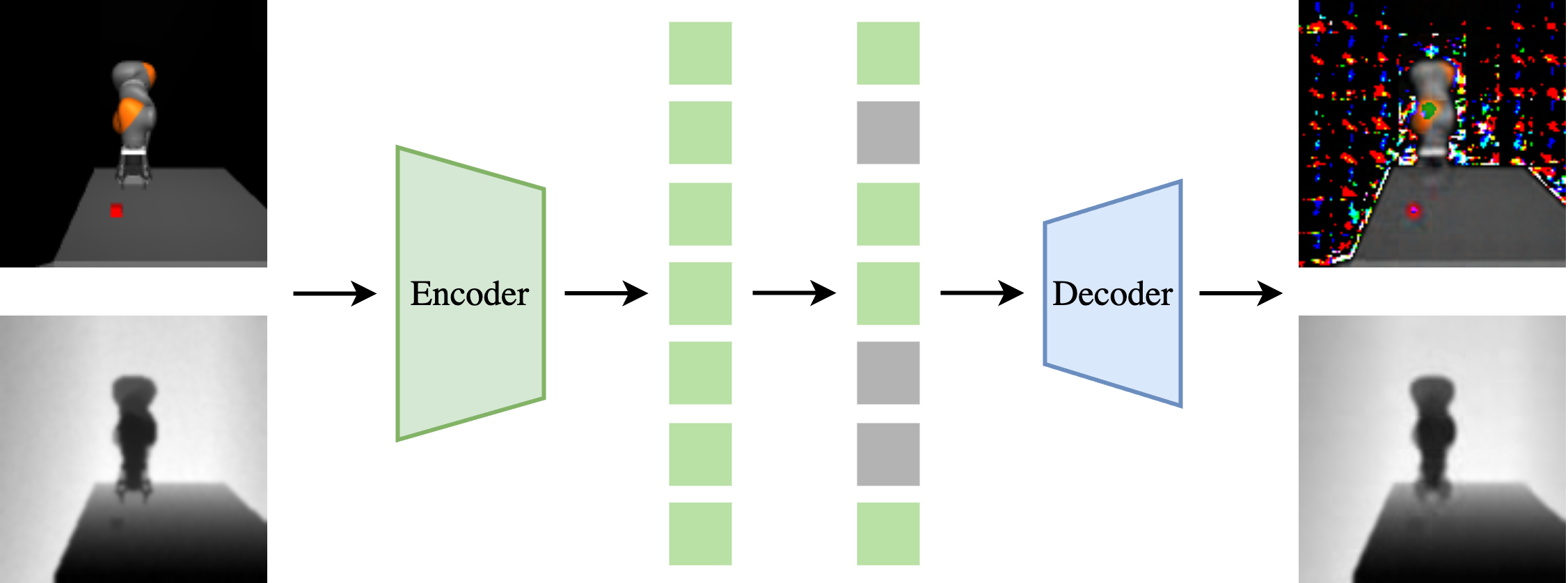}
	\caption{Image reconstruction with the asymmetric decoder.}
	\label{decoder_predicted}
\end{figure}

\textbf{Decoder.} The latent representations of visible patches, along with learnable mask tokens, are fed into a 3-layer, 4-head decoder to reconstruct the original images \cite{he2022masked,sferrazza2024power}. Ablation studies are performed to investigate the effect of decoders on generalization without CURL. The decoder is updated with the following loss:
\begin{equation}
	\mathcal{L}_{dec} = w_{rgb} \cdot \text{MSE} \left(\bm{p}_{rgb}, \hat{\bm{p}}_{rgb}\right) + w_d \cdot \text{MSE} \left(\bm{p}_{d}, \hat{\bm{p}}_{d}\right),
\end{equation}
where $\bm{p}_{rgb}$ and $\bm{p}_{d}$ denote the raw RGB and depth image patches without convolutional processing in Figure~\ref{overview}. $\hat{\bm{p}}_{rgb}$ and $\hat{\bm{p}}_{d}$ are the corresponding reconstructed patches. $w_{rgb} = 1$ and $w_d = 10$ are reconstruction weights for different modalities. $\mathcal{L}_{dec}$ is incorporated into Equation~\ref{critic_loss} and optimized along with the critic. Figure~\ref{decoder_predicted} shows the predicted RGB and depth images at a mask ratio of $50\%$. With the current hyperparameter settings, the decoder can ignore task-irrelevant backgrounds and exhibit potential to improve the generalization compared to the case without CURL, as presented in Table~\ref{ablations}.

\textbf{Mask ratio.} Due to the lack of prior knowledge for identifying task-irrelevant patches, the CURL scheme in this paper only adopts a conservative mask ratio of 0.1. This part will explore whether higher mask ratios contribute to improved generalization. Thus, ablation experiments are conducted with mask ratios of 0, 0.4, and 0.7. As shown in Table~\ref{ablations}, the success rate decreases with increasing mask ratio, accompanied by increased variance in performance. A higher mask ratio results in the random occlusion of more image patches, some of which may contain task-related information. Nevertheless, our results suggest that a minimal level of masking provides essential regularization, whereas excessive patch occlusion significantly degrades generalization. Hence, a CURL scheme with an appropriate mask ratio is beneficial.

\section{Sim2real Experiments} \label{real_exp}
\begin{figure}[th]
	\centering
	\includegraphics[width=.6\textwidth]{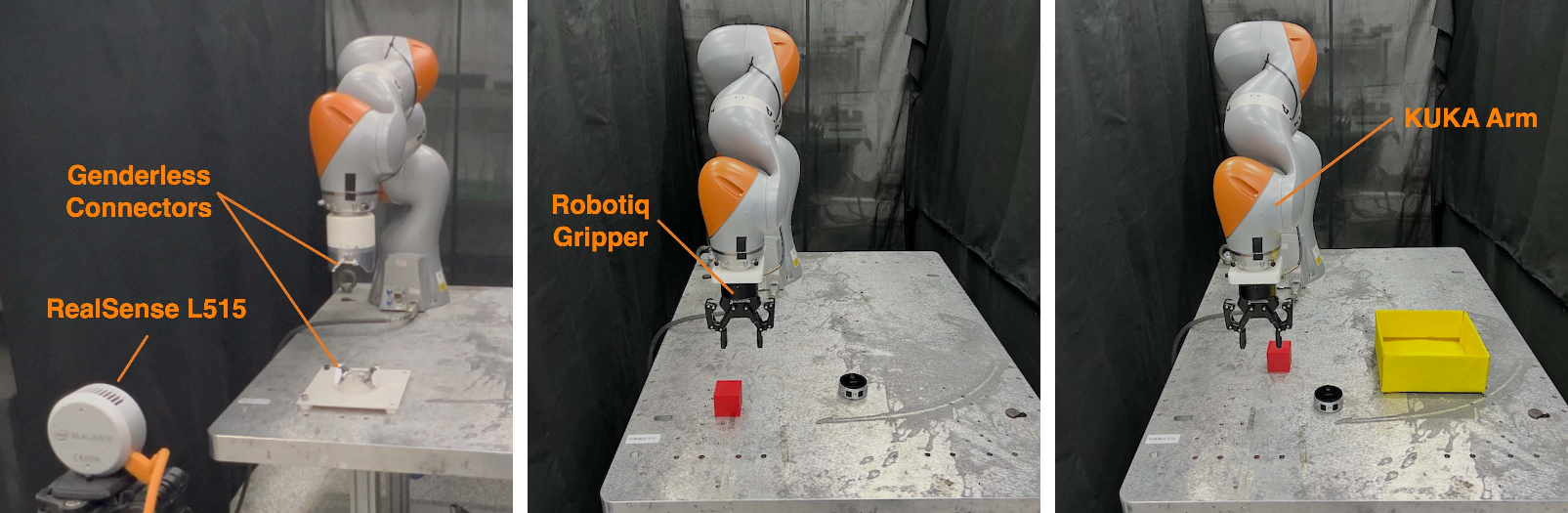}
	\caption{Real-world setup for tasks Assembly, Lift, and PickAndPlace.}
	\label{experiment_setup}
\end{figure}

The KUKA robot controller, PC server, and PC client form the real-world communication network via Socket. The PC server bridges the other two components and delivers visual information and gripper control commands. The trained policy is deployed on the PC client and performs inferences based on the received data. Figure~\ref{experiment_setup} depicts the experimental setup, which mirrors the basic deployment of simulation in Figure~\ref{tasks}. Domain randomization covers scene appearance, camera position, and lighting position. End-effector control is activated for all tasks, and dynamics randomization is excluded \cite{Martin2019Variable}. The training frames are expanded to 1M for PickAndPlace and 800K for the others. The wall-clock training time is approximately 24 hours. The baselines mentioned in Section~\ref{simulations} have been demonstrated to exhibit suboptimal generalization relative to our method and are excluded from further sim2real experiments to mitigate substantial training costs.

\begin{figure*}[t!]
	\centering
	\includegraphics[width=1\textwidth]{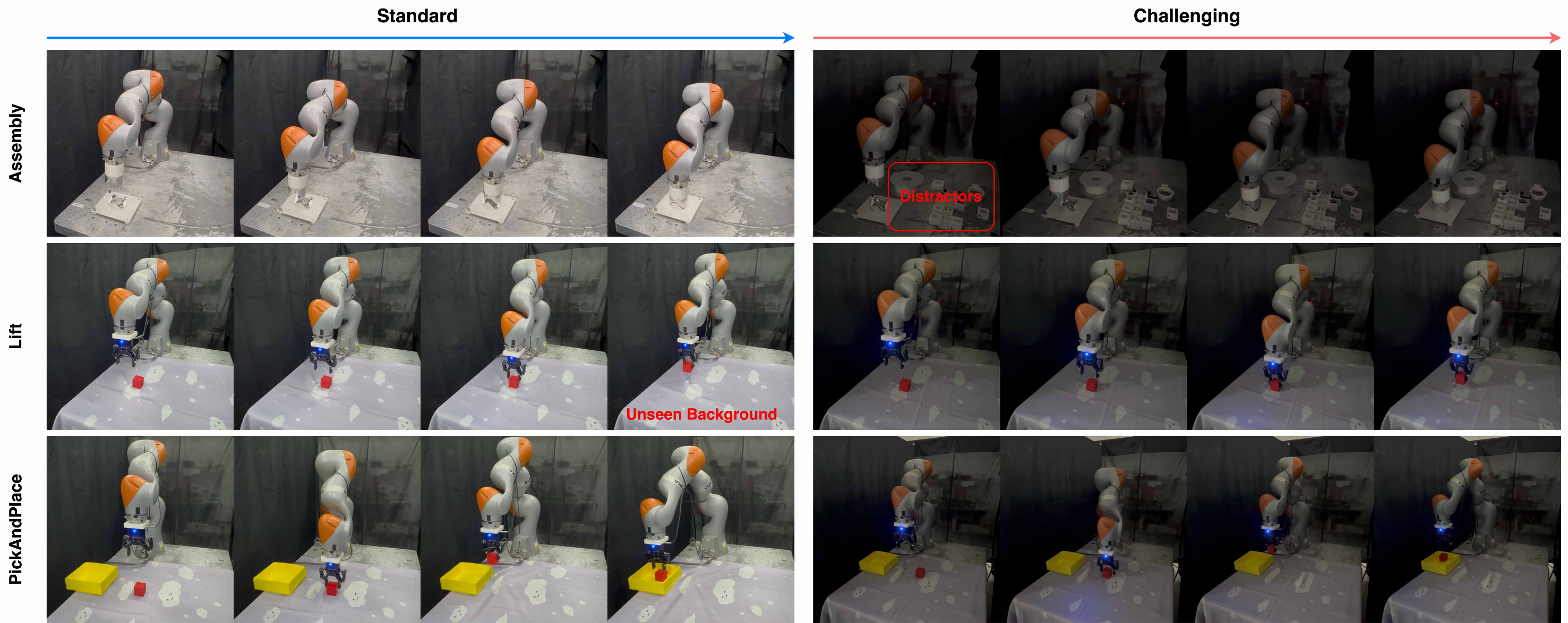}
	\caption{Zero-shot transfer with the proposed visual backbone in standard (left) and challenging (right) real-world scenarios.}
	\label{experiment_process}
\end{figure*}

The trained policy is directly transferred to the experimental platform and evaluated over 30 episodes, half of which involve visually challenging environments. We increase the difficulty of visual observation by adding visual distractors, unseen backgrounds, or dark light compared to the base scenarios. Table~\ref{transfer_performance} summarizes experimental results and demonstrates that our method is effective for zero-shot sim2real transfer. Meanwhile, Figure~\ref{experiment_process} illustrates execution processes in different visual environments using our method. The trained policy enables the KUKA arm to complete the assembly of genderless connectors with variable impedance gains to achieve the desired fitting clearance. The material of the genderless connectors makes it difficult to replicate the reflective visual effect in simulation. Score-CAM visualizations in Table~\ref{real_grad_cam} show that our visual backbone can still capture task-relevant regions when given the observations from challenging scenarios. Meanwhile, the experiment results further confirm that our visual backbone and policy remain robust to discrepancies in material or scene appearance after the training process incorporating data augmentation and curriculum-based domain randomization. For visually challenging scenarios, RGB exhibits large discrepancies compared to those in simulation. Here, our visual backbone can unleash the power of depth information and extract generalizable visual representations to complete the task, which is essential for object-grasping tasks. The experimental video can be found in the submitted supplement.

\begin{table}[t!]
	\centering
	\caption{Sim2Real results with the proposed visual backbone.}
	\label{transfer_performance}
	\begin{tabular}{@{}>{\centering\arraybackslash}m{2.6cm}cc@{}}
	\toprule  
	Task & Standard [\%] & Challenging [\%]\\ 
	\midrule
	Assembly &  $15 / 15$ &  $15 / 15$ \\
	\addlinespace[0.2em]
	Lift &  $ 14 / 15$ &  $ 13 / 15$\\
	\addlinespace[0.2em]
	PickAndPlace &  $ 14 / 15$ &  $ 12 / 15$ \\
	\bottomrule   
	\end{tabular}
\end{table}

\begin{table}[t!]
	\caption{Score-CAM visualization for challenging real-world observations.}
	\centering
	\label{real_grad_cam}
	\begin{tabular}{	
	>{\centering\arraybackslash}m{.7cm}
    >{\centering\arraybackslash}m{2cm}
    >{\centering\arraybackslash}m{2cm}
    >{\centering\arraybackslash}m{2cm}}
	\toprule  
	Task & Assembly & Lift & PickAndPlace \\ 
	\midrule
	RGB & \includegraphics[width=1.8cm]{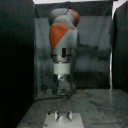} &  \includegraphics[width=1.8cm]{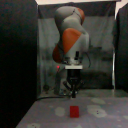} & \includegraphics[width=1.8cm]{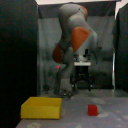}\\
	Depth & \includegraphics[width=1.8cm]{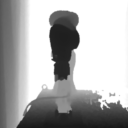} & \includegraphics[width=1.8cm]{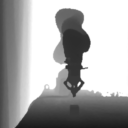} & \includegraphics[width=1.8cm]{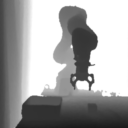} \\
	Score-CAM & \includegraphics[width=1.8cm]{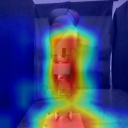}  & \includegraphics[width=1.8cm]{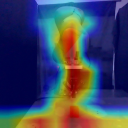} & \includegraphics[width=1.8cm]{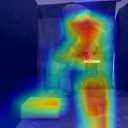} \\
	\bottomrule   
	\end{tabular}
\end{table}

\begin{figure}[th]
	\centering
	\includegraphics[width=.6\textwidth]{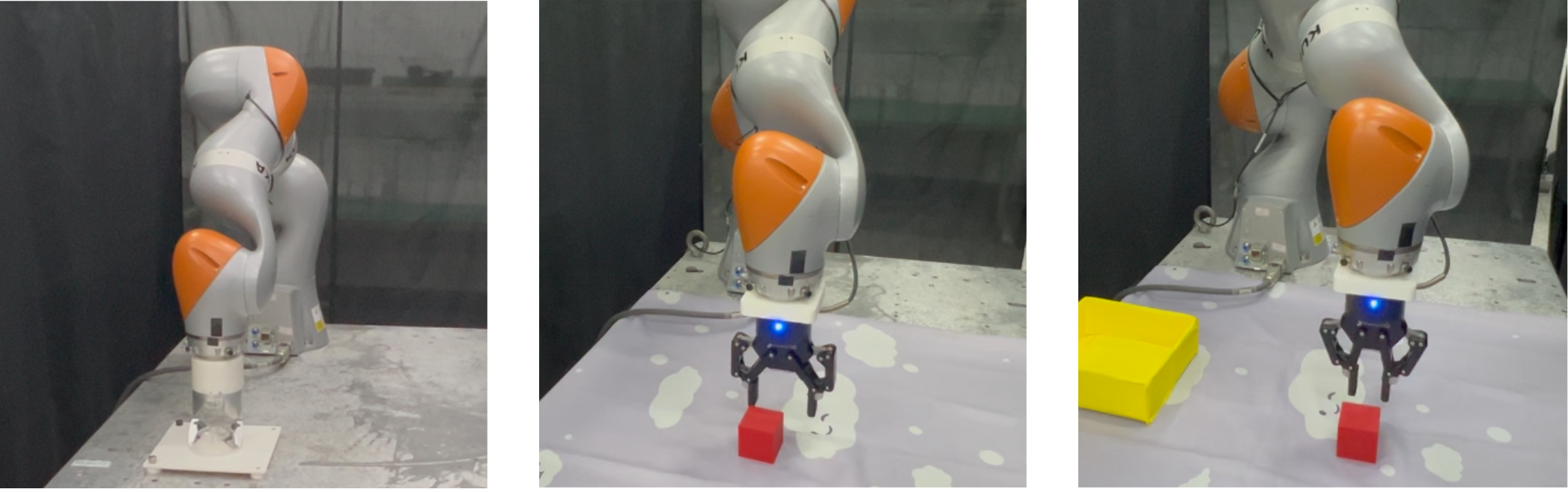}
	\caption{Failed sim2real transfer scenarios with only RGB input.}
	\label{rgb_failed_process}
\end{figure}

Besides, depth ablation experiments are conducted to investigate the sim2real transfer performance of policies without depth input. The CNN stem for depth and the modality embedding in Figure~\ref{overview} are removed. With the same hyperparameters, the single-modal policies that rely solely on RGB information fail to demonstrate as favorable generalization as the aforementioned multimodal policies, which are summarized in Table~\ref{rgb_transfer_performance}. As shown in Figure~\ref{rgb_failed_process}, the poor performance of single-modal policies is explained by the difficulties in perceiving the relative position between the target and the end-effector. Thus, even minor sim2real gap can induce significant decay in generalization of single-modal policies. The real-world ablations demonstrate that depth information is essential for enhancing the success rate of sim2real transfer and accomplishing manipulation tasks.

\begin{table}[th]
	\centering
	\caption{Sim2Real results with single-modal policies.}
	\label{rgb_transfer_performance}
	\begin{tabular}{@{}>{\centering\arraybackslash}m{2.2cm}cc@{}}
	\toprule  
	Task & Standard [\%] & Challenging [\%] \\ 
	\midrule
	Assembly &  $3 / 15$ &  $0 / 15$ \\
	\addlinespace[0.2em]
	Lift &  $ 1 / 15$ &  $ 0 / 15$\\
	\addlinespace[0.2em]
	PickAndPlace &  $ 0 / 15$ &  $ 0 / 15$ \\
	\bottomrule   
	\end{tabular}
\end{table}

Finally, the critics obtained after domain randomization should ideally provide similar $Q$-value estimations for the original and noisy observations, which leads the policy to output similar actions. The embedding features of the penultimate layer in the $Q$-value network are projected onto the 2D space for dimensionality reduction and then visualized via t-SNE. 1K random observations in simulation are sampled to form two distinct state-action pairs, in which noisy RGB observations are augmented through random overlay. As shown in Figure~\ref{experiment_process}, the trained critics can distinguish different observations and demonstrate distinct clustering ability, which contributes to mitigating perception discrepancies. Thus, the generalization of the actor can also be enhanced as training proceeds. 

\begin{figure}[th]
	\centering
	\includegraphics[scale=.4]{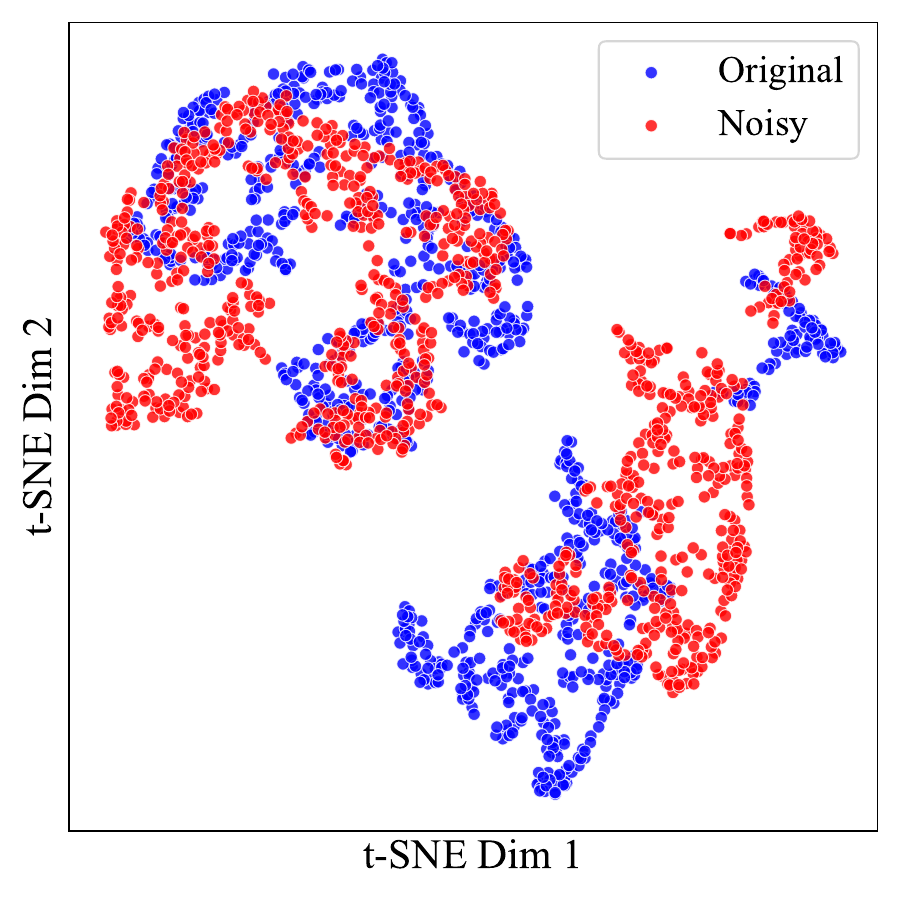}
	\caption{t-SNE visualization of $Q$-value embeddings for original and noisy observations processed with random overlay.}
	\label{tsne}
\end{figure}

\section{Conclusion And Limitations}
This paper focuses on exploring the impact of depth modality and visual backbone architectures on generalization and sim2real performance of visual RL policies. A visual backbone integrating ViT and CNN is first presented to fuse RGB and depth modalities. A contrastive unsupervised learning scheme is then constructed to extract consistent information from masked and unmasked tokens processed by ViT. The simulation results show that the fusion of depth information has a direct impact on the generalization of visual RL policies across various visual backbones, and our proposed backbone demonstrates better performance. A curriculum-based domain randomization is developed for sim2real transfer. Real-world experiments indicate that our policy can achieve zero-shot sim2real transfer and adapt to diverse visual scenarios.

The early and late fusion baselines in this paper are constructed based on common CNN architectures. However, numerous ConvNets remain available for comparison. This work demonstrates that ViT-based backbones are promising alternatives to some ConvNets in visual RL and sim2real applications. We are also dedicated to exploring the potential of transformer-like architectures, which is consistent with many prior studies in computer vision \cite{xiao2021early, liu2021swin}. Besides, further work will dive into the generalization to different camera viewpoints. 

\section{Appendix}

\begin{table}[th]
	\vspace{-0.3cm}
	\centering
	\caption{Hyperparameters used for training}
	\begin{tabular}{@{}p{5cm}@{}p{6.3cm}@{}}
	\toprule
	Hyperparameter    & Value     \\ 
	\hline
	Image size        & 128 $\times$ 128 \\
	Frame stack & 3 (RGB), 1 (Depth)   \\
	Discount factor $\gamma$    & 0.99            \\
	Replay Buffer size          & 1e6            \\
	Batch size & 256 \\
	$N$-step return  &  3 \\
	Optimizer & Adam   \\
	Learning rate & 1e-4 \\
	Max episode steps & 500 (PickAndPlace), 400 (Others) \\
	\multirow{2}{*}{Exploration schedule }  &  Linear $\left(1.0, 0.1, 100\text{K} \right)$ (Others)\\
											&  Linear $\left(1.0, 0.1, 300\text{K} \right)$ (PickAndPlace)\\ 
	Action repeat  & 2  \\
	CNN encoder & Conv (c=[32, 64, 128, 256]) \\ 
	Actor \& Critic & Linear (c=[256, 1024, 1024]) \\
	Patch size of ViT & 16  \\
	Heads of ViT & 8 \\
	Embedding dim of ViT & 512 \\
	Head dim of ViT & 32 \\
	Layers of ViT & 4 \\
	Mask ratio for CURL & 0.1 \\
	\hline
	\end{tabular}
	\label{common_hyper}
\end{table}


\bibliographystyle{IEEEtran}

\end{document}